\documentclass[10pt,twocolumn,letterpaper]{article}

\usepackage{cvpr}              % To produce the CAMERA-READY version
\usepackage{algorithm}
\usepackage{algpseudocode}
\usepackage{amsmath}
\usepackage{graphicx}
\usepackage{amsmath}
\usepackage{amssymb}
\usepackage{booktabs}
\usepackage{multirow}
\usepackage[pagebackref,breaklinks,colorlinks]{hyperref}

\usepackage[capitalize]{cleveref}
\crefname{section}{Sec.}{Secs.}
\Crefname{section}{Section}{Sections}
\Crefname{table}{Table}{Tables}
\crefname{table}{Tab.}{Tabs.}

\def\cvprPaperID{9252} % *** Enter the CVPR Paper ID here
\def\confName{CVPR}
\def\confYear{2023}

\begin{document}

%%%%%%%%% TITLE - PLEASE UPDATE
\title{Balancing Privacy Protection and Interpretability in Federated Learning}

\author{Zhe Li \and Honglong Chen \and Zhichen Ni \and Huajie Shao}
% China University of Petroleum
% Institution1 address\\
% {\tt\small firstauthor@i1.org}
% For a paper whose authors are all at the same institution,
% omit the following lines up until the closing ``}''.
% Additional authors and addresses can be added with ``\and'',
% just like the second author.
% To save space, use either the email address or home page, not both
% Institution2\\
% First line of institution2 address\\
% {\tt\small secondauthor@i2.org}

\maketitle

%%%%%%%%% ABSTRACT
\begin{abstract}
Federated learning (FL) aims to collaboratively train the global model in a distributed manner by sharing the model parameters from local clients to a central server, thereby potentially protecting users' private information. Nevertheless, recent studies have illustrated that FL still suffers from information leakage as adversaries try to recover the training data by analyzing shared parameters from local clients. To deal with this issue, differential privacy (DP) is adopted to add noise to the gradients of local models before aggregation. It, however, results in the poor performance of gradient-based interpretability methods, since some weights capturing the salient region in feature map will be perturbed. To overcome this problem, we propose a simple yet effective adaptive differential privacy (ADP) mechanism that selectively adds noisy perturbations to the gradients of client models in FL. We also theoretically analyze the impact of gradient perturbation on the model interpretability. Finally, extensive experiments on both IID and Non-IID data demonstrate that the proposed ADP can achieve a good trade-off between privacy and interpretability in FL. 
% Importantly, it is robust to the gradient leakage attack.
\end{abstract}

%%%%%%%%% BODY TEXT
\section{Introduction} \label{sec:intro}
% This paper proposes a novel adaptive differential privacy (ADP) approach to achieve a good trade-off between privacy protection and interpretability in federated learning (FL). 

\begin{figure}[!tb]
  \centering\includegraphics[width=3.26in,height=2.3in]{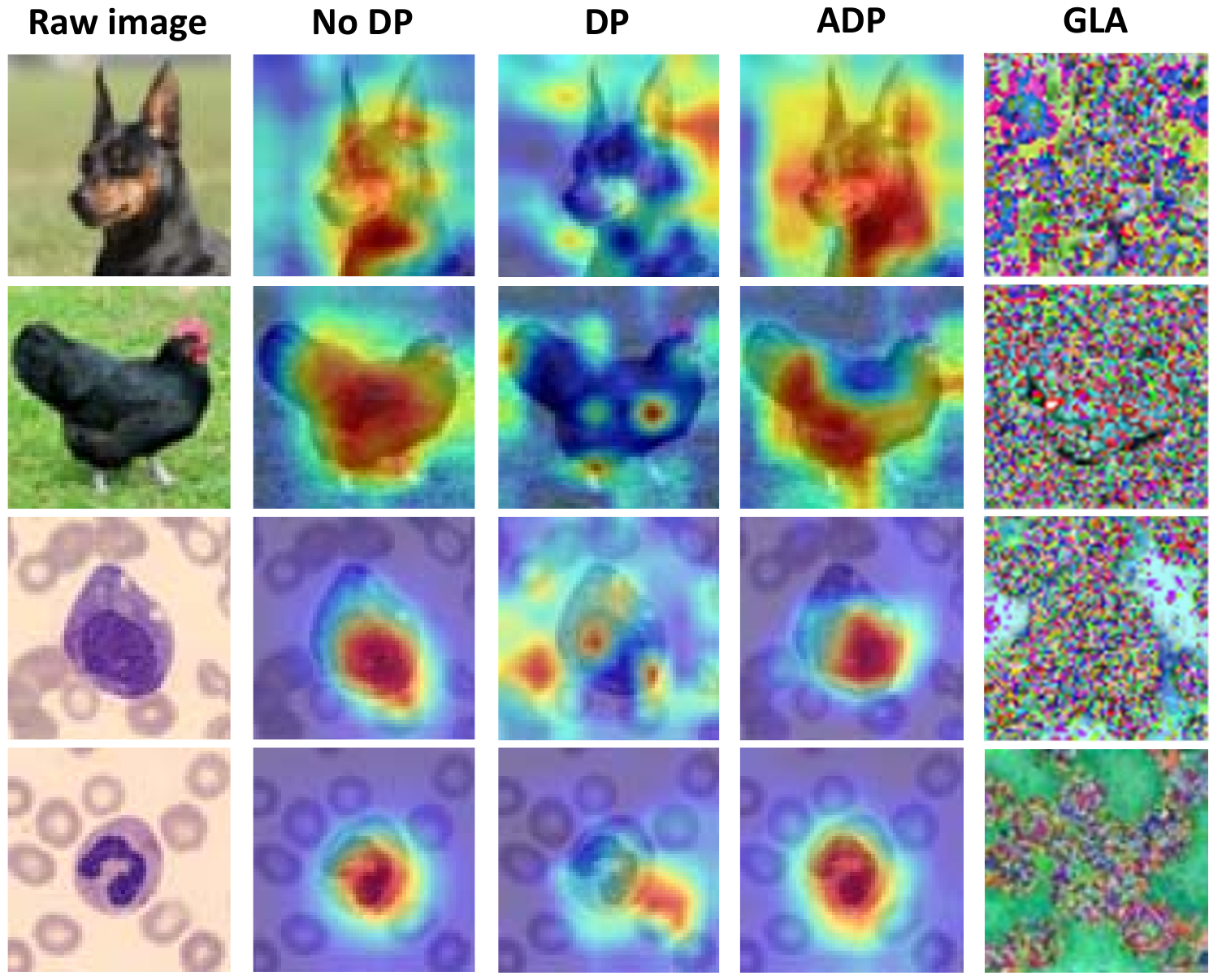}
  \caption{Visualizations of different methods for model inerpretability using Grad-CAM~\cite{Selvaraju2017}. Red regions denote higher scores for classification. We can see that the regular DP will lead to poor model interpretability while our ADP can still retain the performance of interpretability. More importantly, it can still be robust to gradient leakage attack and guarantee privacy protection.}
  \label{fig:demo_example}
\end{figure}

Federated learning (FL)~\cite{McMahan2017} is an emerging privacy-preserving mechanism that collaboratively trains a global model without exchanging data with different clients~\cite{Lin2020ensemble,Li2021model,Kairouz2021,Deng2022TPDS}. Since it has the potential to protect the private information, FL has been widely applied to a variety of application domains, such as healthcare~\cite{Rieke2020,Dou2021federated}, insurance industry~\cite{Liu2020AAAI}, and Internet of Things (IoTs)~\cite{Zhang2021JSAC,Wang2022JSAC,Ghimire2022}.
Recent works have pointed out that FL is vulnerable to gradient leakage attacks (GLA) that try to reconstruct the training data from the publicly shared gradients with a central server~\cite{Zhu2019deep,Wang2022protect}. To deal with this problem, one of commonly used defense strategies is differential privacy (DP)~\cite{Dwork2014}, which injects noise to the model parameters (weights or gradients) before they are uploaded to a central server~\cite{Yin2021comprehensive}. However, DP protects private information of users at the cost of model accuracy in FL. Hence, some methods~\cite{} have been developed to achieve a good trade-off between privacy preservation and accuracy. Existing works mainly pay attention to accuracy but neglect the model interpretability in FL. In reality, injecting noise to model parameters using differential privacy will hurt the gradient-based interpretability, as illustrated in Figure \ref{fig:demo_example}. As a result, the lack of interpretability will hinder the deployment of FL in safety-critical domains, such as healthcare and autonomous driving. While there exist very few studies~\cite{Patel2022,Harder2020} on the trade-offs between interpretability and privacy protection in general machine learning (not federated learning), they mainly focus on \textit{perturbation-based interpretation} by modifying the input features. To our best knowledge, none of prior works has explored the trade-off between privacy protection and \textit{gradient-based interpretability in FL}. 
%%%
In this paper, we will study how to trade off privacy protection and interpretability in the context of privacy-preserving federated learning. The key challenge lies in how to smartly add noise to the model parameters while trying to retain its interpretability. To address this challenge, we first theoretically analyze the impact of injected noise on gradient-based model interpretability. This work uses the well-known Grad-CAM~\cite{Selvaraju2017} as a representative. Based on the analysis results, we develop a novel adaptive differential privacy (ADP) approach that selectively adds noisy perturbations to the weights of local clients. More specifically, we do not add noise to the important (large) weights that capture the salient regions of feature map but inject noise to small weights that may not affect the model interpretability too much. Figure \ref{fig:demo_example} shows an illustration example of the proposed ADP for ensuring privacy protection and interpretability. Finally, we evaluate the performance of our ADP on multiple benchmark datasets. Evaluation results demonstrate that our approach is simple yet effective, which can improve the model interpretability while guaranteeing privacy in FL. Importantly, it is robust to gradient leakage attack.

% We illustrate a motivating example of interpretability in Figure \ref{fig:demo_example} to clearly demonstrate the superiority of our proposed ADP over DP in improving interpretability. Not only that, ADP is also resistant to GLA, demonstrating its ability to provide sufficient privacy protection.

In sum,  the \textbf{main contributions} of this work include: 1) to our best knowledge, we are the first to explore the trade-offs between privacy protection and model interpretability in federated learning; 2) we provide theoretical analysis on the impact of injected noise on gradient-based interpretability methods; 3) a novel adaptive differential privacy (ADP) approach is proposed to inject noise to model parameters smartly; and 4) experimental results demonstrate the proposed approach can achieve a good trade-off between privacy protection and model interpretability in FL.
\section{Preliminaries}
In this section, we introduce some basic knowledge about gradient-based interpretability, federated learning, and differential privacy that will be used in the paper.
\subsection{Grad-CAM}\label{sec:grad}
As a typical gradient-based interpretability approach, Gradient-weighted Class Activation Mapping (Grad-CAM)~\cite{Selvaraju2017} aims to provide visual explanations on the tasks of computer vision by using the gradients to capture the salient regions of feature map in the last convolutional layer of deep neural networks.

In this paper, we use image classification as a running example to introduce the basic idea of Grad-CAM. Let $y^c$ be the prediction score (before softmax activation) of class $c$ and $Z^k$ denote the feature map of a certain (e.g., last) convolutional layer. The gradient of $y^c$ with respect to $Z^k$ is denoted by $\frac{\partial y^c}{\partial Z^k}$. Then we can leverage the gradients via back-propagation to get the following neuron importance weights as:
\begin{equation}
\begin{aligned}
    \alpha_k^c = \frac{1}{M}\sum_i \sum_j \frac{\partial y^c}{\partial Z_{ij}^k}, \nonumber
\end{aligned}
\end{equation}
where $k$ represents the $k$-th channel of feature map $Z$, $M$ represents the product of the width and height of the feature map, $\alpha_k^c$ denotes the importance of feature map $Z^k$ for class $c$, and $Z_{ij}^k$ is the data of feature map $Z$ at the coordinate $(i, j)$ in channel $k$.

Next, we can use weighed sum of forward activation maps and the ReLU activation function to obtain the localization map Grad-CAM $L_{Grad-CAM}^c$ as follows.
\begin{equation}
\begin{aligned}
   L_{Grad-CAM}^c = ReLU\Big(\sum_k \alpha_k^c Z^k\Big).  \nonumber
\end{aligned}
\end{equation}
Note that, ReLU is used to remain features that have a positive impact on class $c$. 

% \textcolor{red}{Gradient-weighted Class Activation Mapping (Grad-CAM), a high-resolution class-discriminative localization technique that creates visual explanations from any CNN-based network, leverages gradient information flowing into the last convolutional layer of the CNN to determine the significance of each neuron for the decision of interest~\cite{Selvaraju2017}.}

% We utilize the image classification task in this paper as an example to demonstrate the computational process of GradCAM. The network prediction $y^c$ (before softmax activation) of class $c$ and feature maps $A^k$ of a convolutional layer are first obtained by forward propagation of the neural network. The activation map of the last convolutional layer is frequently used due to its rich high-level semantic information and detailed spatial information. The gradient of $y^c$ with respect to $A^k$ is computed as $\frac{\partial y^c}{\partial A^k}$. Then by back-propagation, \textcolor{red}{the gradients are global-average-pooled to obtain the neuron importance weights}:
% \begin{equation}
% \begin{aligned}
%     \alpha_k^c = \frac{1}{Z}\sum_i \sum_j \frac{\partial y^c}{\partial A_{ij}^k},
% \end{aligned}
% \end{equation}
% where $k$ represents the $k$-th channel of feature map $A$, $Z$ represents the product of the width and height of the feature map, $\alpha_k^c$ denotes the importance of feature map $k$ for class $c$, and $A_{ij}^k$ is the data of feature map $A$ at the coordinate $(i, j)$ in channel $k$.

\subsection{Federated learning}\label{sec:federated}
% \shao{double check if is aligned with your Algorithm 2}
Federated learning is an effective privacy-preserving technique that collaboratively trains a global model without sharing data with each client. The detailed implementation of FL can be summarized as follows. First, a central server sends a global model with weight initialization $w_0$ to local clients. Then $K$ selected clients will train the model with local data, and then each of them, e.g., client $k$, will upload the model parameter $w_k^t$ to the central server at the $t$-th round of communication. On server side, it will update $w_t$ by aggregating the weights from selected clients, yielding 
\begin{equation}\label{}
w_{t} \leftarrow \text{Aggregate}(w_1^t,\ldots,w_k^t,\ldots, w_K^t).   \nonumber
\end{equation}
After that, the server will broadcast the global parameter to local clients for the next round of training. In local training, the weights of each client can be updated with
\begin{equation}\label{}
w_k^{t+1} \leftarrow w_{t} - \eta \nabla F_k(w_k^{t+1}),  \nonumber
\end{equation}
where $\eta$ is the learning rate and $\nabla F_k(w_k^{t+1})$ is the gradient of client $k$.

\subsection{Differential privacy}
As a privacy-preserving method, differential privacy (DP) attempts to maximize the useful information retrieved from a database while minimizing the expenditure of privacy resources~\cite{Adnan2022}. In recent years, DP has been widely used in federated learning to defend against gradient leakage attack. One of the most popular DP techniques is called ($\epsilon$, $\delta$)-differential privacy~\cite{Dwork2014}, given by
\begin{equation}
\begin{aligned}
      {\rm Pr}\big[M(D_i)\in S \big] \leq e^\epsilon \, {\rm Pr}\big[M(D^{'}_i)\in S \big] + \delta, \nonumber
\end{aligned}
\end{equation}
where $D_i$ and $D^{'}_i$ are two neighboring datasets and only one single entry of data differs between them. $\epsilon$ is the privacy budget. The smaller $\epsilon$, the higher privacy protection. In addition, the parameter $\delta$ denotes the upper bound of the output probability that can be changed by adding or removing a record in the dataset, i.e., the privacy protection standard. 

% denotes the upper bound of the output probability that can be changed by adding or removing a record in the dataset, i.e., the privacy protection standard. The smaller the value, the stricter the standard, hence it is also known as privacy budget. Additionally, the parameter $\delta$ is typically set to a small constant, which denotes \textcolor{red}{the event that the probability ratio for two adjacent datasets $D_i$ and $D^{'}_i$ not being constrained by $e^ \epsilon$ after a privacy-preserving mechanism} has been applied~\cite{Wei2020}. 
In the applications of federated learning, we often achieve differential privacy by adding a moderate amount of Gaussian noise and clipping the gradient. 
% Differential privacy applied to federated learning is usually achieved by clipping the gradient and adding a moderate amount of Gaussian noise. \textcolor{red}{To ensure that the Gaussian noise distribution $n \sim N(0, \sigma^2)$ satisfies ($\epsilon$, $\delta$)-DP, it is necessary to choose noise scale $\sigma \geq c\Delta s/ \epsilon$, and the constant $c \geq \sqrt{2 {\rm ln}(1.25/\delta)}$ for $\epsilon \in (0, 1)$}. Additionally, $\Delta s$ represents \textcolor{red}{the sensitivity of a real-valued function} $s$ given by $\Delta s = {\rm max}_{D_i, D^{'}_i}\big|\big|s(D_i)-s(D^{'}_i)\big|\big|$.

\section{Effect of DP on interpretability}
In this section, we provide theoretical analysis about the impact of DP on gradient-based interpretability in FL.

We take an $L$-layer deep neural network (DNN) where the last two layers are composed of one convolutional layer (the second to last layer) and one fully connected layer as an example. The detailed network architecture is given by
\begin{align*} 
& Z^{(1)}  = W^{(1)}X  + b^{(1)}, \\
& H^{(1)} = \sigma\big(Z^{(1)}\big),\\
& \ \ \ \ \ \ \ \ \ \ \ \vdots\\
& Z^{(l-1)} = Conv\big(H^{(l-2)},W^{(l-1)}\big) + b^{(l-1)}, \\
& H^{(l-1)} = \sigma\big(Z^{(l-1)}\big), \\
& Y =  W^{(l)} H^{(l-1)} + b^{(l)},
\end{align*}
where $Y$ is the output logits before softmax layer and $\sigma(\cdot)$ is the activation function. In addition, $Conv()$ is the second to last convoluational layer after $L-2$ fully connected layers, and the last layer is fully connected.
% \widetilde{W}

% We firmly believe that there is a trade-off between privacy protection of DP and model interpretability. The more noise is introduced to the model, the stronger the privacy protection is, which accordingly, reduces its accuracy and interpretability. 
In the following, we will investigate how the injected noise added to the model parameters influences the gradient-based interpretability in privacy-preserving FL. In this work, we mainly focus on the widely used GradCAM for model interpretability. 
% In this paper, we investigate the trade-off between privacy protection and interpretability applying the widely used gradient-based algorithm GradCAM as a technique to demonstrate visual interpretability. We analyze the trade-off between the noise $n_\epsilon$ added to parameters by applying DP and the gradients $\frac{\partial y^c}{\partial A^k}$ calculated in GradCAM by deriving the forward and backward propagation of this network.

\subsection{Feedforward propagation}
Since the GradCAM approach mainly relies on the feature map of the last convolutional layer for model interpretability, for simplicity, we will focus on the last two layers of the DNN, including one convolutional layer and one fully connected layer. Specifically, let $H^{(l-2)}$ and $W^{(l-1)}$ denote the input of the convolutional layer and the convolution kernel, respectively. Then we add the independent identical distribution (IID) Gaussian noise, $n_\epsilon \sim N(0, \zeta^2)$, to the convolution kernel $W^{(l-1)}$ and the last fully connected layer, yielding
\begin{align*}
& \widetilde{W}^{(l-1)} =  W^{(l-1)} + n_\epsilon, \\
& \widetilde{W}^{(l)} =  W^{(l)} + n_\epsilon.
\end{align*}

% Without loss of generality, it can be extended to more layers using the same strategy below. 

% The input of the convolutional layer is $H^{(l-2)}$, the convolution kernel is $W^{(l-1)}$. After adding Gaussian noise with distribution $n_\epsilon \sim N(0, \zeta^2)$, the convolution kernel is represented as $\widetilde{W}^{(l-1)} =  W^{(l-1)} + n_\epsilon$, and the weight matrix is denoted as $\widetilde{W}^{(l)}$.

After injecting noise, the last two layers of the DNN can be written as:
\begin{align*} 
& Z^{(l-1)} = \text{Conv}\big(H^{(l-2)},\widetilde{W}^{(l-1)}\big) + b^{(l-1)}, \\
& H^{(l-1)} = \sigma\big(Z^{(l-1)}\big), \\
& Y = \widetilde{W}^{(l)}H^{(l-1)} + b^{(l)}.
\end{align*}
Note that for simplicity, we do not show the injected noise to the bias term in the derivation, but will \textit{add the noise to all the parameters in the DNN.} Next, we still study how the noise impacts the weights during backpropagation. 

\subsection{Backpropagation}
Assume that the loss function is denoted as $L$ and the prediction (before softmax) for class $c$ as $Y_c$. To obtain the gradient of $Y_c$ with respect to the feature map in the $i$-th channel $\frac{\partial Y_c}{\partial Z_i^{(l-1)}}$, we need to compute the backpropagation in the following steps.
\begin{align*}\footnotesize
& \frac{\partial Y_c}{\partial H_i^{(l-1)}} =  \widetilde{W}_{c,i}^{(l)}, \\
& \frac{\partial H_i^{(l-1)}}{\partial Z_i^{(l-1)}} = \sigma'(Z_i^{(l-1)}).
\end{align*}

Based on the above derivation, we can obtain the following expression for $ \frac{\partial Y_c}{\partial Z_i^{(l-1)}}$:
\begin{small}\begin{align*}\label{eq:back_CNN}
    & \frac{\partial Y_c}{\partial Z_i^{(l-1)}} = \frac{\partial Y_c}{\partial H_i^{(l-1)}}\frac{\partial H_i^{(l-1)}}{\partial Z_i^{(l-1)}} = \widetilde{W}_{c,i}^{(l)}\sigma'(Z_i^{(l-1)}) \nonumber \\ 
    % = & \widetilde{W}^{(l)}\cdot \sigma'\Big(\text{Conv}\big(H^{(l-2)},\widetilde{W}^{(l-1)}\big) + b^{(l-1)}\Big) \\
    & = \big(W_{c,i}^{(l)} + n_\epsilon \big) \sigma'\Big(\text{Conv}_i\Big(H^{(l-2)},\big(W^{(l-1)} + n_\epsilon \big)\Big) + b_i^{(l-1)}\Big).
\end{align*}
\end{small}
The above equation is composed of two terms: the first term is the derivative of the last fully connected layer and the second term is the derivative of convolutional layer. Next, we briefly analyze the influence of noise on the feature map using gradient based interpretability, as introduced in Section~\ref{sec:grad}. We can see that if we only add noise to the fully connected layer, it will affect the feature map used for model interpretability. While if we only inject noise to the convolutional layer, it will change the output of activation function, thereby changing the feature map.

The above simple analysis suggests that the regular DP will affect the gradient-based interpretability in federated learning. It motivates us to develop a new differential privacy method to improve the model interpretability in the following section. 
\section{Proposed method}\label{sec:model}
In this section, we are going to design a new adaptive differential privacy (ADP) mechanism that selectively injects noise into the model parameters to improve the model interpretability while retaining privacy protection in federated learning. 
% Differential privacy based federated learning approaches contribute to improving the level of privacy protection of target models, however their interpretability tends to gradually diminish as privacy protection is enhanced. Therefore, we believe that there is tradeoff between privacy protection and interpretability, and conduct a fundamentally theoretical analysis in this section to support our idea. Furthermore, in order to improve this tradeoff, we propose ADP, a new adaptive differential privacy (ADP) mechanism that selectively exempts a certain percentage of the Gaussian noise added to the parameters that are most crucial for making interpretations, thus enhancing the interpretable performance as much as possible with adequate privacy protection, which is vital for deep learning models, particularly in healthcare domain.

\subsection{DP-based federated learning}
We first briefly present the core idea of DP-based Federated Learning. As described in Section~\ref{sec:federated}, federated learning collaboratively trains a global model in a central server by aggregating the weights or gradients from selected clients and then sends them back to the clients after aggregation. Different from the regular FL, DP-based federated learning attempts to add Gaussian noise into the weights before aggregation. Then the weights with noisy perturbation in local clients will be uploaded to the central server. As discussed above, injecting noise will negatively affect the gradient-based interpretability. As a result, the lack of model interpretability hinders the widespread deployment of FL in safety-critical applications, such as healthcare and autonomous driving. 

\subsection{Adaptive differential privacy mechanism}
To deal with the challenge above, we will explore how to improve model interpretability while preserving privacy in the context of federated learning. As we know, gradient-based interpretability, such as Grad-CAM, will target the salient regions (high weights) in the feature map to interpret the important pixel-level features. The key insight is that when the noise with negative value is injected to the large weights, it will counteract the neuron important weights, leading to poor interpretability of FL. Conversely, if we add large positive noise to less important weights, the insignificant regions become salient ones, misleading the interpretability of prediction results. Inspired by this, we propose to design a simple yet effective adaptive differential privacy (ADP) mechanism that selectively injects noise into model parameters. The main idea is that we only inject noise into the less important weights, such that their values should not be over the neuron important weights. To achieve this, we need to sort the weights in a descending order, and then add noise to the bottom $r$\% weights, $\widetilde{W}_{r}$. If $\widetilde{W}_{r}$ exceeds the smallest value of the top $r$\% weights, it should be clipped. 

% we do not add noise to the important weights and
% motivated by protecting the regions in the feature map that play a key role in improving interpretability by exempting them from some degree of noise. We therefore propose an adaptive differential privacy mechanism that selectively adds noise perturbations to specific parameters to mitigate the adverse effects of privacy protection on model interpretability.
\subsection{Application of ADP to FL}
We apply the proposed ADP to improve the interpretability of DNN in federated learning. In this work, we mainly focus on improving the trade-off between privacy and interpretability of image classification using ResNet~\cite{he2016deep} consisting of multiple convolutional layers. In general, the shallow layers of ResNet contain high-resolution features but with low-level semantic information. The deep layers extract some more complex and abstract features with high-level semantics, but at the same time the feature resolution decreases. Considering both ``shallow local features" and ``deep global features", we choose to selectively add noise to the weights of the last convolutional layer of each block of ResNet (8 parameters for ResNet18). We will conduct experiments to validate the good performance of the proposed strategy. 

\begin{algorithm}[h]
  \caption{Adaptive differential privacy mechanism} \label{alg:adp}
  \label{alg::conjugateGradient}
  \begin{algorithmic}[1] %% add line number
  \Require $r\in (0, 1)$: ratio of noise addition, $\epsilon$: privacy budget of differential privacy,
  $\delta$: relaxation item of differential privacy, $\Delta s$: sensitivity of differential privacy,
  $target\_layers$: selected convolutional layers, $local\_parameters$: local parameters trained on clients, $target\_parameters$: parameters of $target\_layers$.
  \Ensure Local parameters with adaptive noise.

    \For{each $layer$ in $target\_layers$}
    \State $W_{cam}$ $ \leftarrow$ GradCAM(target\_layer=$layer$)
    \State $m$ $ \leftarrow$  multiply $r$ with number of elements in $W_{cam}$
    \State $e$ $ \leftarrow$  sort $W_{cam}$ and take out the $(m-1)$-th element 
        \For{each weight $w$ in $W_{cam}$}
        \If {$w \ge e$} 
        \State Set $w$ to 0;
        \Else 
        \State Set $w$ to 1;
        \EndIf
        \State Noise mask $M$ $ \leftarrow$ $W_{cam}$;
        \EndFor
    \EndFor
     \For{each $p$ in $local\_parameters$}
     \State $N$ $ \leftarrow$ Gaussian\_Simple($\epsilon$, $\delta$, $\Delta s$);
        \For{each $q$ in $target\_parameters$}
        \If {$p = q$} 
        \State $N$ $\leftarrow$ $N * M$;
        \EndIf
    \EndFor
     \State $p$ $ \leftarrow$ $p$ + $N$;
     \EndFor
  \end{algorithmic}
\end{algorithm}

The proposed ADP mechanism is summarized in Algorithm \ref{alg:adp}. The names of the selected convolutional layers and the parameter of each layer are stored in ``target layers" and ``target parameters", respectively. Line 2 shows the weights of feature maps of each convolutional layer obtained by GradCAM. Line 4 sorts the weights of each target layer and Lines 14-21 selectively add noise to the less important weights for interpretability. In order to easily add noise to the weights, we create a mask matrix that can generate binary value (0 and 1) by setting a threshold to judge whether perturbing weights or not. This mask matrix can help adaptively add noise to the model parameters in ``target parameters". Note that, the proposed mechanism is able to flexibly choose the ratio of injected noise to the weights in each target layer. It is simple, easy to use, yet effective to achieve a good trade-off between interpretability and privacy. 

% set the noise addition ratio in the range of $(0, 1)$ arbitrarily, the network layers and their parameters for adaptive noise addition can also be selected according to different requirements.

\section{Evaluation}\label{sec:experiment}
\renewcommand\arraystretch{1.3}   %调整表格的行距
\begin{table*}[!t]
\caption{Accuracy comparison of different methods on IID data using 3 random seeds.}
\label{table1}
\begin{center}
\begin{tabular}{c|c|c|ccc|ccc}
\toprule[1.25pt]
\multirow{2}{*}{Dataset} & No DP     & DP     & \multicolumn{3}{c|}{Random DP}    & \multicolumn{3}{c}{ADP}    \\ \cline{2-9}
& 0\% noise & 100\% noise & \multicolumn{1}{c|}{75\% noise} & \multicolumn{1}{c|}{50\% noise} & 25\% noise & \multicolumn{1}{c|}{75\% noise} & \multicolumn{1}{c|}{50\% noise} & 25\% noise \\
\midrule[0.75pt]
MNIST   & 97.71\%   & 86.08\%     & \multicolumn{1}{c|}{86.93\%}    & \multicolumn{1}{c|}{89.77\%}    & 91.39\%    & \multicolumn{1}{c|}{\textbf{87.81\%}} & \multicolumn{1}{c|}{\textbf{89.90\%}} & \textbf{91.77\%} \\ \hline
Blood   & 92.00\%   & 50.77\%     & \multicolumn{1}{c|}{55.47\%}    & \multicolumn{1}{c|}{60.97\%}    & 62.62\%    & \multicolumn{1}{c|}{\textbf{65.81\%}} & \multicolumn{1}{c|}{\textbf{67.01\%}} & \textbf{69.67\%}   \\
\hline
Animals & 69.92\%   & 53.10\%     & \multicolumn{1}{c|}{54.14\%}    & \multicolumn{1}{c|}{56.55\%}    & 58.43\%    & \multicolumn{1}{c|}{\textbf{54.56\%}} & \multicolumn{1}{c|}{\textbf{57.04\%}} & \textbf{59.07\%} \\ 
\bottomrule[1.25pt]
\end{tabular}
\end{center}
\end{table*}

\renewcommand\arraystretch{1.3}   %调整表格的行距
\begin{table*}[htb]
\caption{Accuracy comparison of different methods on Non-IID data using 3 random seeds.}
\vspace{-0.1in}
\tabcolsep 0.08in
\label{table2}
\begin{center}
\begin{tabular}{c|c|c|ccc|ccc}
\toprule[1.25pt]
\multirow{2}{*}{Dataset} & No DP     & DP     & \multicolumn{3}{c|}{Random DP}    & \multicolumn{3}{c}{ADP}    \\ \cline{2-9}
& 0\% noise & 100\% noise & \multicolumn{1}{c|}{75\% noise} & \multicolumn{1}{c|}{50\% noise} & 25\% noise & \multicolumn{1}{c|}{75\% noise} & \multicolumn{1}{c|}{50\% noise} & 25\% noise \\
\midrule[0.75pt]
MNIST   & 96.96\%   & 71.45\%     & \multicolumn{1}{c|}{74.10\%}    & \multicolumn{1}{c|}{75.04\%}    & 81.82\%    & \multicolumn{1}{c|}{\textbf{74.34\%}} & \multicolumn{1}{c|}{\textbf{77.63\%}} & \textbf{82.19\%} \\ \hline
Blood   & 89.25\%   & 45.36\%     & \multicolumn{1}{c|}{50.20\%}    & \multicolumn{1}{c|}{52.08\%}    & 54.82\%    & \multicolumn{1}{c|}{\textbf{50.73\%}} & \multicolumn{1}{c|}{\textbf{54.42\%}} & \textbf{57.04\%}   \\
\hline
Animals & 58.10\%   & 39.02\%     & \multicolumn{1}{c|}{40.87\%}    & \multicolumn{1}{c|}{41.74\%}    & 44.90\%    & \multicolumn{1}{c|}{\textbf{41.27\%}} & \multicolumn{1}{c|}{\textbf{42.62\%}} & \textbf{45.48\%} \\ 
\bottomrule[1.25pt]
\end{tabular}
\end{center}
\end{table*}

In this section, we carry out extensive experiments to evaluate the performance of proposed ADP in FL. First of all, we study the effect of ADP on classification accuracy. Then we exlore the impact of ADP on gradient-based interpretability. Finally, we leverage gradient leakage attack to measure the robustness of the proposed method.
\subsection{Datasets}
We evaluate the proposed ADP on three benchmark datasets, MNIST~\cite{MNIST}, Blood~\cite{Blood}, and Animals~\cite{Animal}, as described below.

\textbf{MNIST}. This dataset contains a total of 70,000 grayscale images of size 28×28, divided into a training set of 60,000 images (and labels) and a test set of 10,000 images (and labels). 

% It contains a total of ten categories from numbers 0 to 9.

\textbf{Blood}. This dataset contains 8-class categories of microscopic peripheral blood cell images. It contains a total of $17092$ RGB images of individual blood cells, which are split into 13600 and 3492 images for training and testing. In our experiment, we resize the images from the original 3 × 360 × 363 pixels to 3 × 64 × 64 pixels.

\textbf{Animals}. This dataset contains 10 animal species, commonly used in the field of image classification. Then we split the total 26,000 images into 20,800 and 5,200 for training and testing. To keep the same image size, we uniformly resize them to 3 × 64 × 64 pixels.

% We performed a preliminary data cleaning on the original dataset to remove a few erroneous images, and 

\subsection{Model configuration and parameter settings}
We use ResNet18 with 17 convolutional layers and 1 fully connected layer for image classification, and choose SGD as the optimizer with a learning rate of 0.001. We set a total of 100 clients for federated training, with 0.1-fraction clients chosen in each communication round. We set the number of communication rounds to 1000, and the local training batch to 100. At each round, we set the epoch to 1 for MNIST and Blood, and 5 for Animals. We implement experiments on both IID and Non-IID scenarios. 

In the setting of differential privacy, we follow the common setting of $\delta$ to $10^{-5}$, test the effect of $\epsilon$ on accuracy by setting $\epsilon$ to 100, 10, 5, 1, 0.1, respectively and the effect of gradient clipping parameter $c$ on accuracy by setting $c$ to 5, 10, 20, 40, 80, respectively. Note that the parameter $\epsilon$ is the total privacy budget, while each client's budget in each communication round is the total privacy budget divided by the product of the fraction and the number of communication rounds.

\subsection{Baselines}
Since we are the first to study the trade-off between privacy and gradient-based interpretability, it is hard to find appropriate benchmark baselines in the prior works. Nevertheless, we compare the proposed ADP method with the following baselines

\textbf{No DP}. ``No DP" means that no differential privacy is applied to the federated averaging model, which will serve as an essential baseline to provide a comparison for DP-based models.

\textbf{DP}. ``DP" stands for applying differential privacy on the basis of No DP, which adds randomly 100\% of noise into the model parameters. 

% Since tests on the effect of different parameters on accuracy were performed in advance, we chose $c=5$, and $\epsilon=1$ for MNIST, $\epsilon=5$ for Blood and Animals, as the settings for With DP method. The reasons for the choice of parameters are described in \ref{section 5.4}.

\textbf{Random DP}. For a fair comparison, we choose a baseline called ``Random DP", which will add the same ratio of noise to the weights as the proposed ADP method. The only difference between this method and ADP is that it randomly adds noise to the weights while ADP selectively adds noises to the less important weights.

% \textbf{Adaptive DP}. ``Adaptive DP" refers to the adaptive differential privacy mechanism proposed in this paper, which is capable of adaptively adding noise of arbitrary ratio to specific parameters, and we set the ratio as 0.25, 0.5, and 0.75 for experiments.

\begin{figure}[t]
\centering
\subfloat[MNIST]{
\includegraphics[width=3.3in]{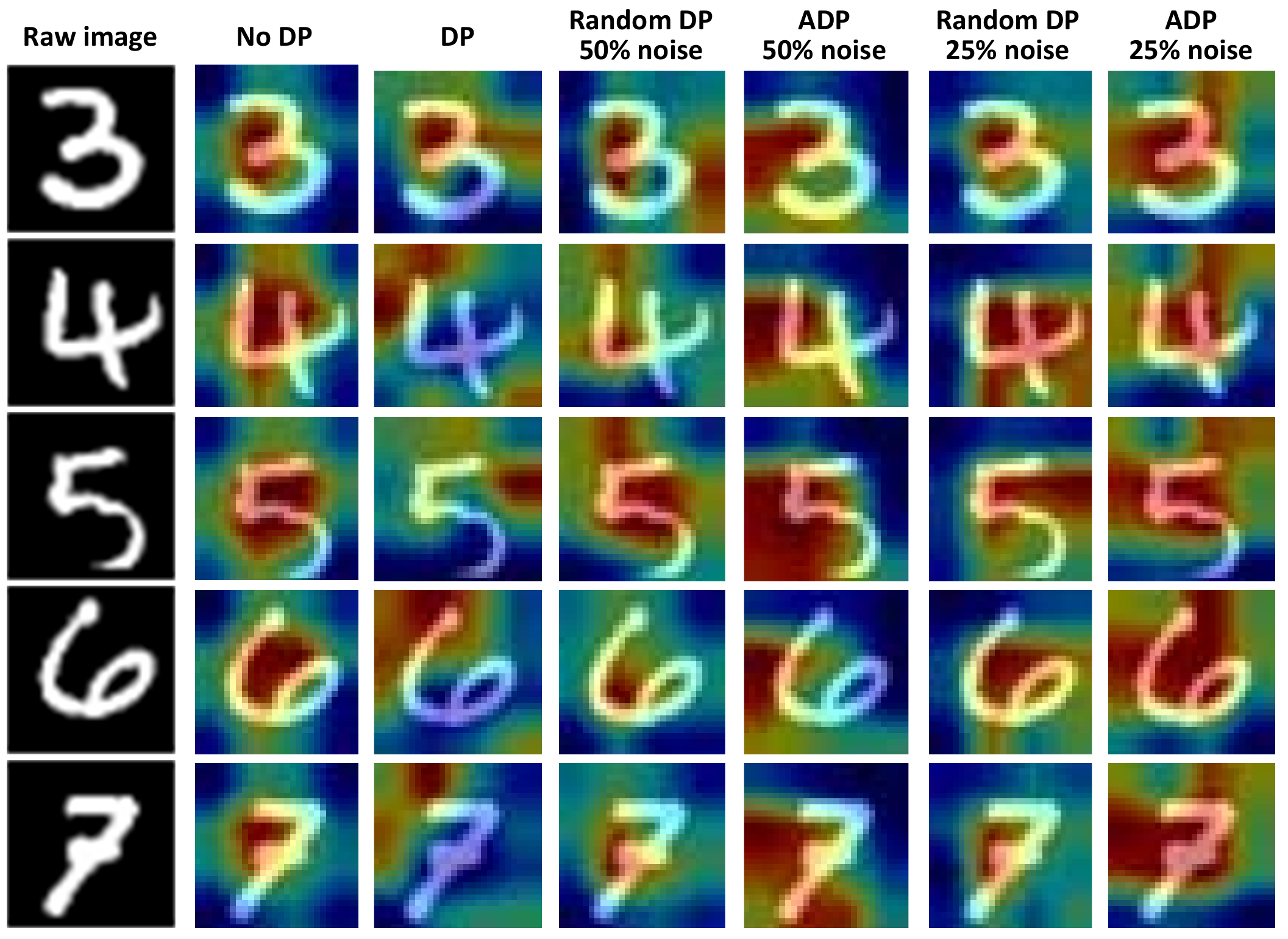}} \\ 
\subfloat[Blood]{
\includegraphics[width=3.3in]{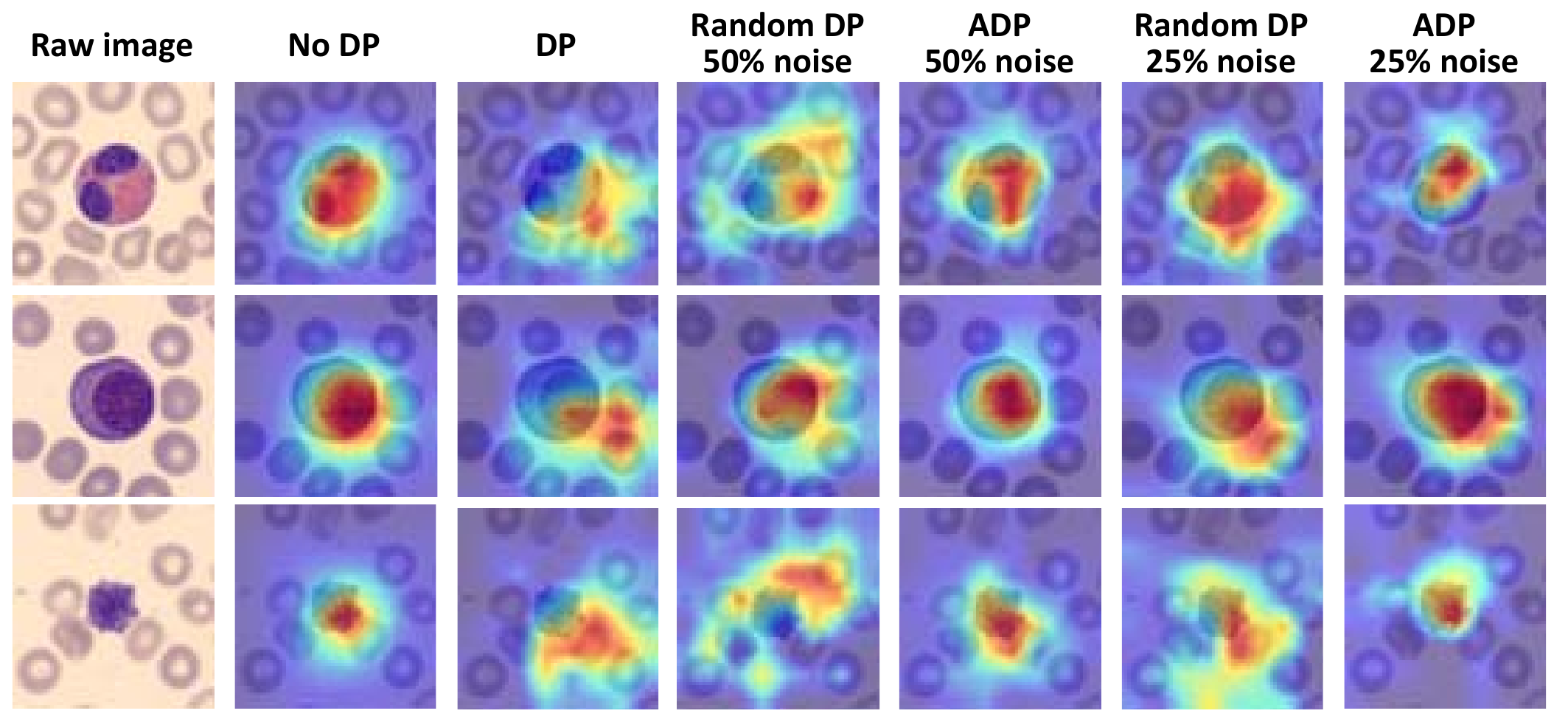}} \\
\subfloat[Animals]{
\includegraphics[width=3.3in]{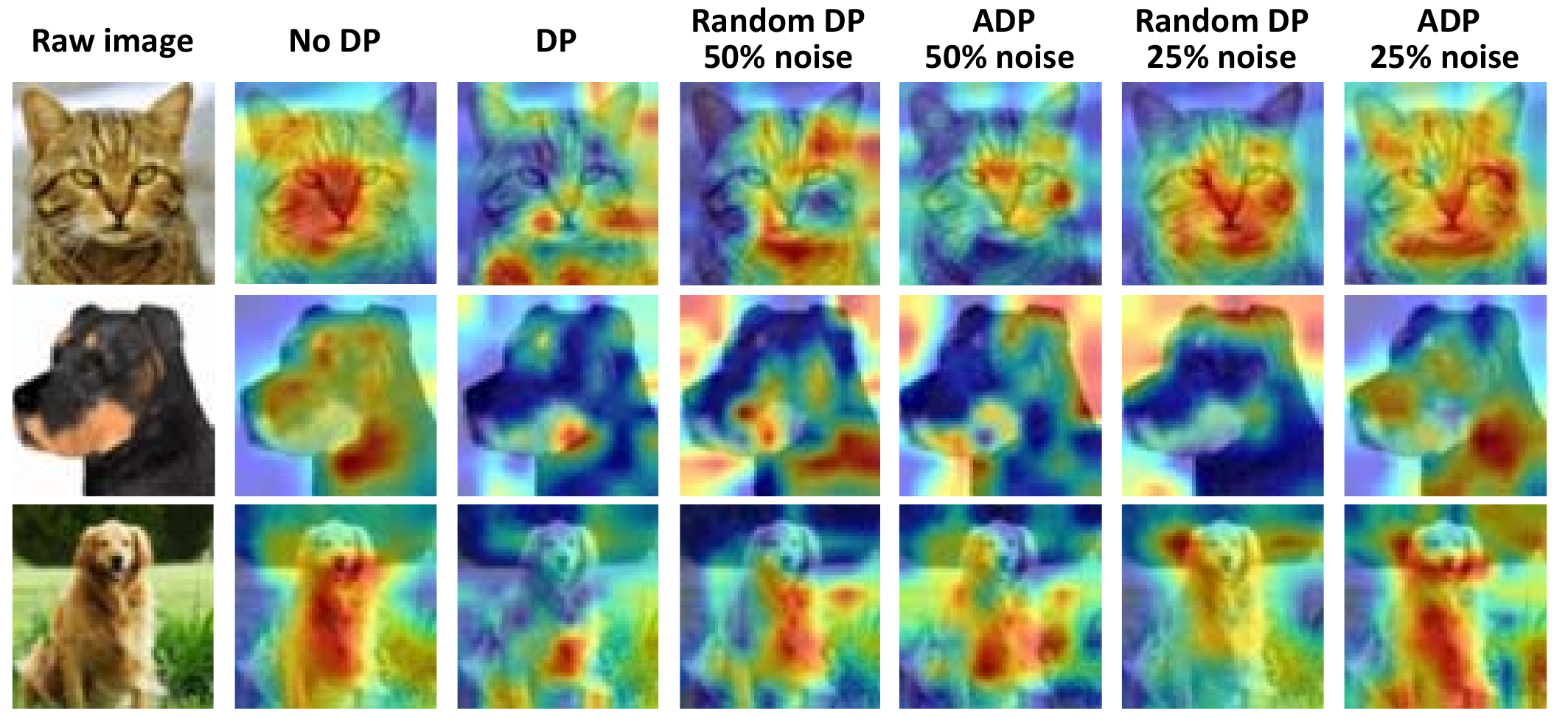}} \\
\caption{Explainable visualizations of different methods using heatmap under IID scenario. Red regions denote higher scores for classification.}\label{fig:ex-iid}
\end{figure}

\begin{figure}[t]
  \centering
  \subfloat[MNIST]{
  \includegraphics[width=3.3in]{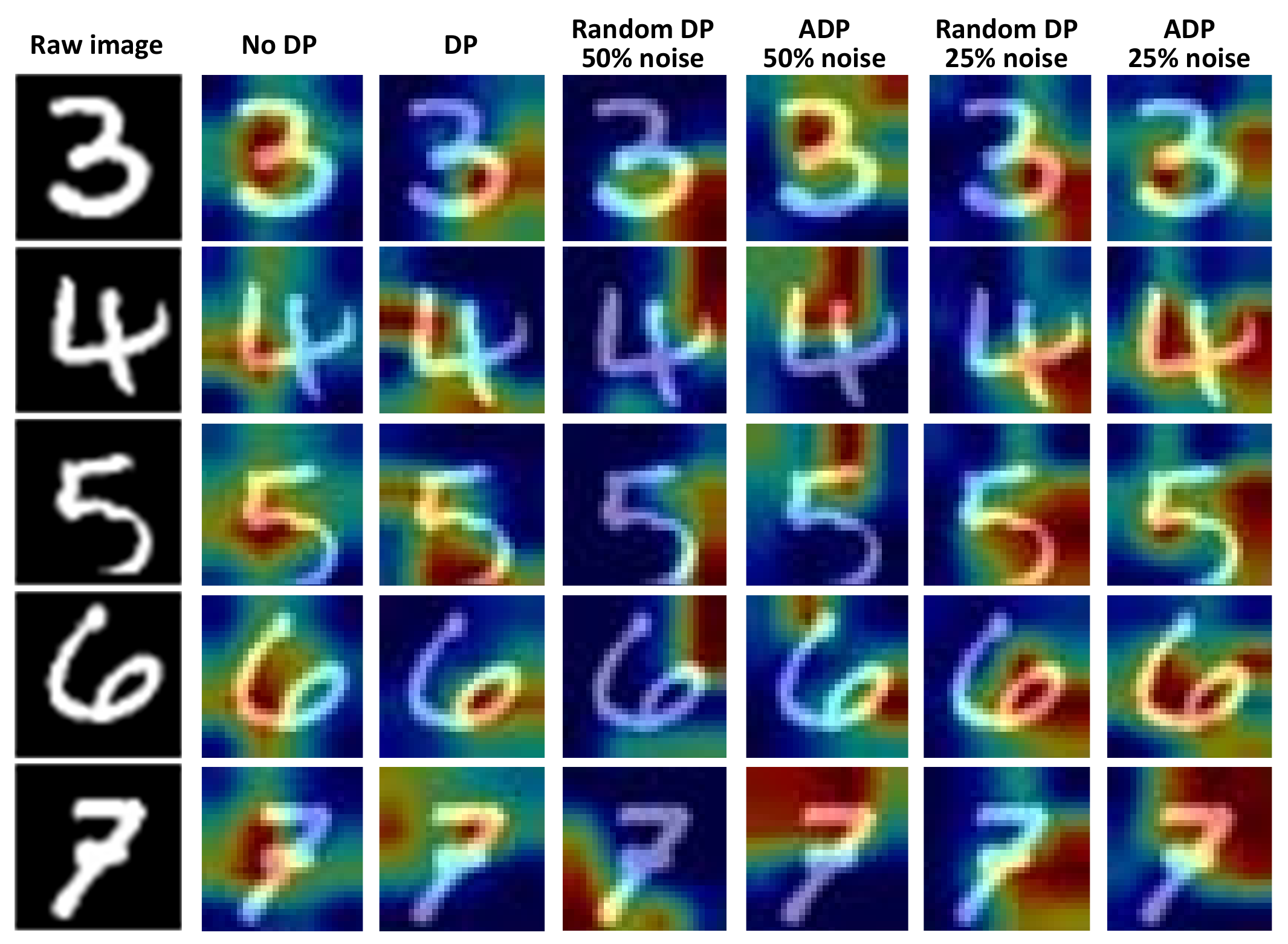}} \\
  \subfloat[Blood]{
  \includegraphics[width=3.3in]{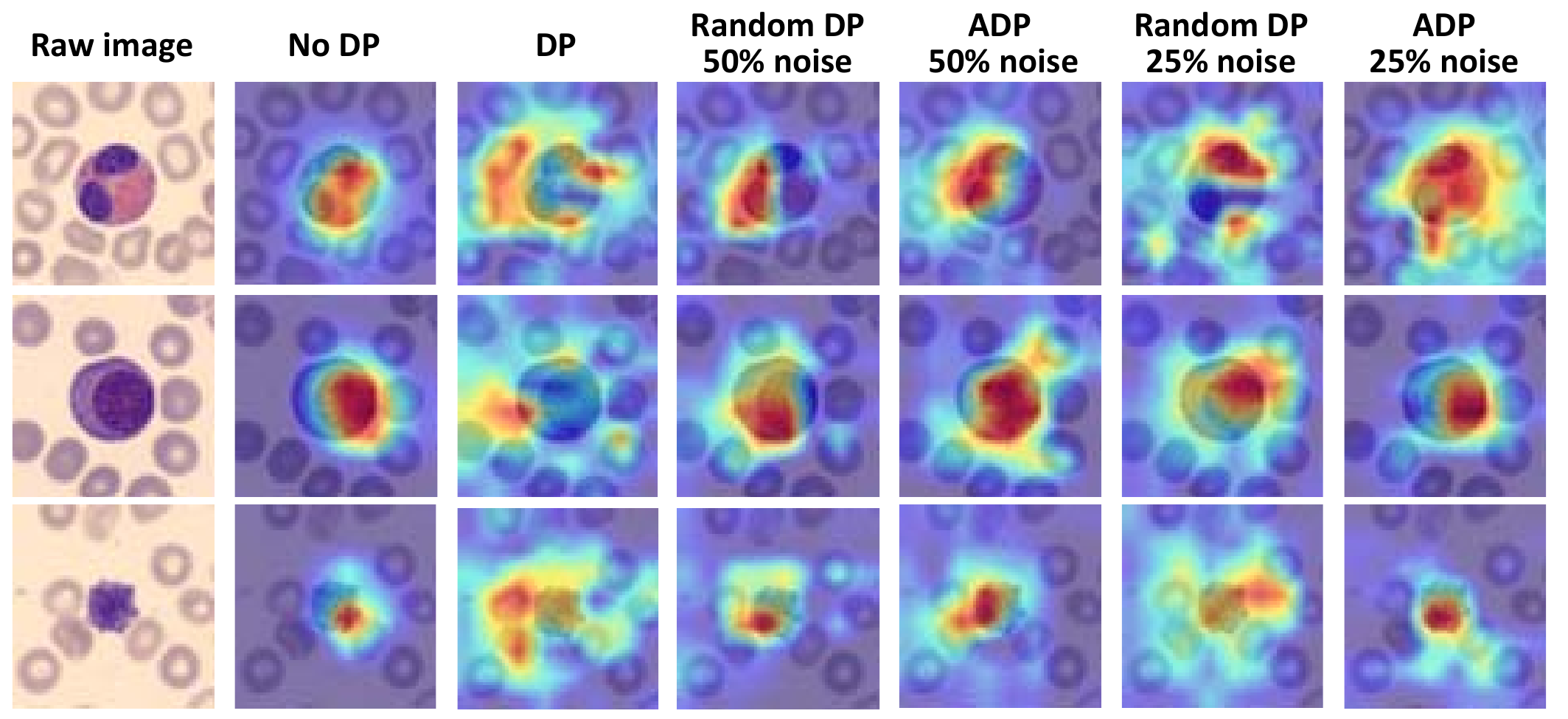}} \\
  \subfloat[Animals]{
  % \label{fig:subfig:c}
  \includegraphics[width=3.3in]{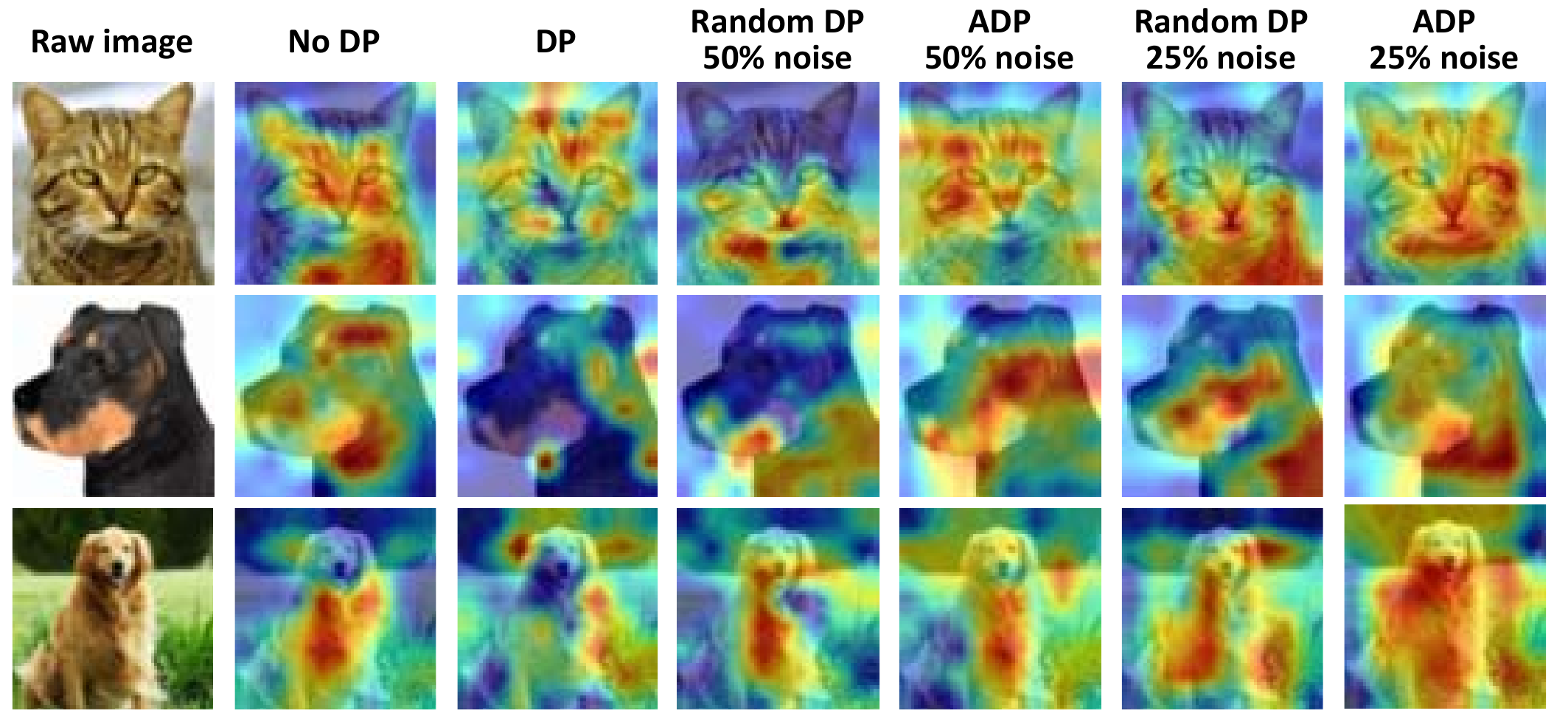}} \\
  \caption{Explainable visualizations of different methods using heatmap under Non-IID scenario. Red regions denote higher scores for classification.}\label{fig:non-iid}
  \end{figure}

\subsection{Impact of ADP on accuracy}\label{section 5.4}

Before studying the impact of ADP on model interpretability, we first would like to check whether the proposed method can improve accuracy. In this experiment, we choose to inject 25\%, 50\%, and 75\% noise to model parameters of Random DP and ADP. In addition, we observe that accuracy decreases as the clipping parameter $c$ rises. In order to not affect the accuracy too much, we set parameter $c$ to 5. Regarding privacy budget $\epsilon$, we set it to $1$, $5$, and $5$ for MNIST and the other two datasets by considering the trade-off between accuracy and privacy protection. We compare our methods to the baselines under the IID and Non-IID scenarios. 

\begin{figure*}[t]
  \centering\includegraphics[width=1\textwidth]{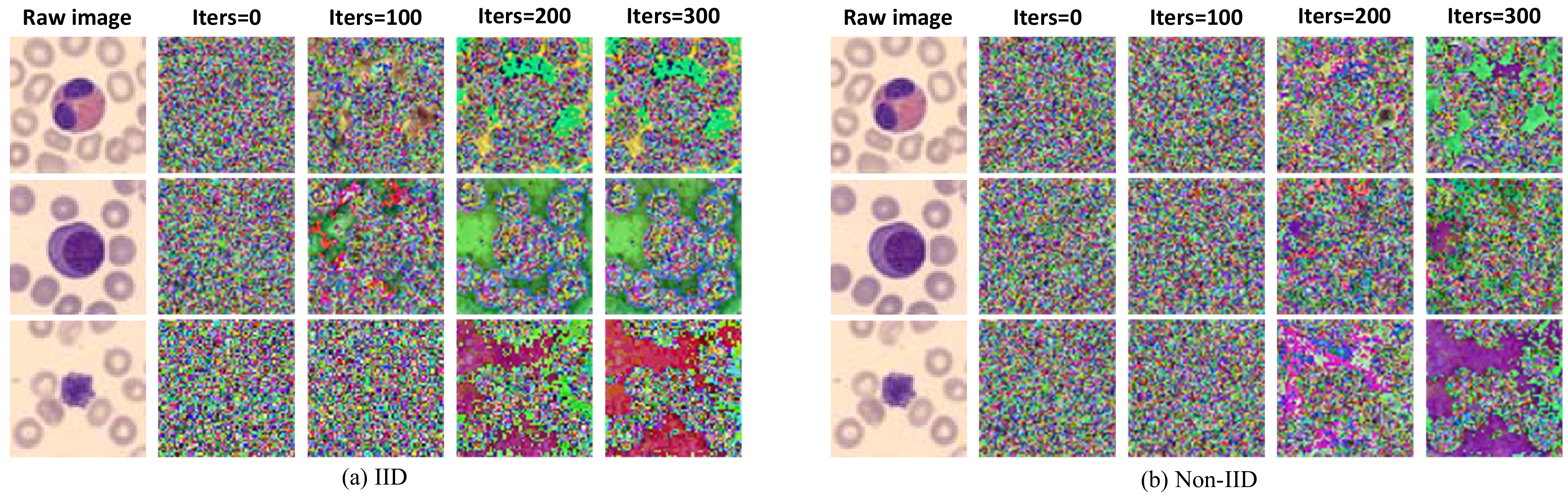}
  \caption{DLG attack against ADP with ratio of 0.25 on Blood.}
  \label{fig:attackblood}
\end{figure*}

\begin{figure*}[!tb]
  \centering\includegraphics[width=1\textwidth]{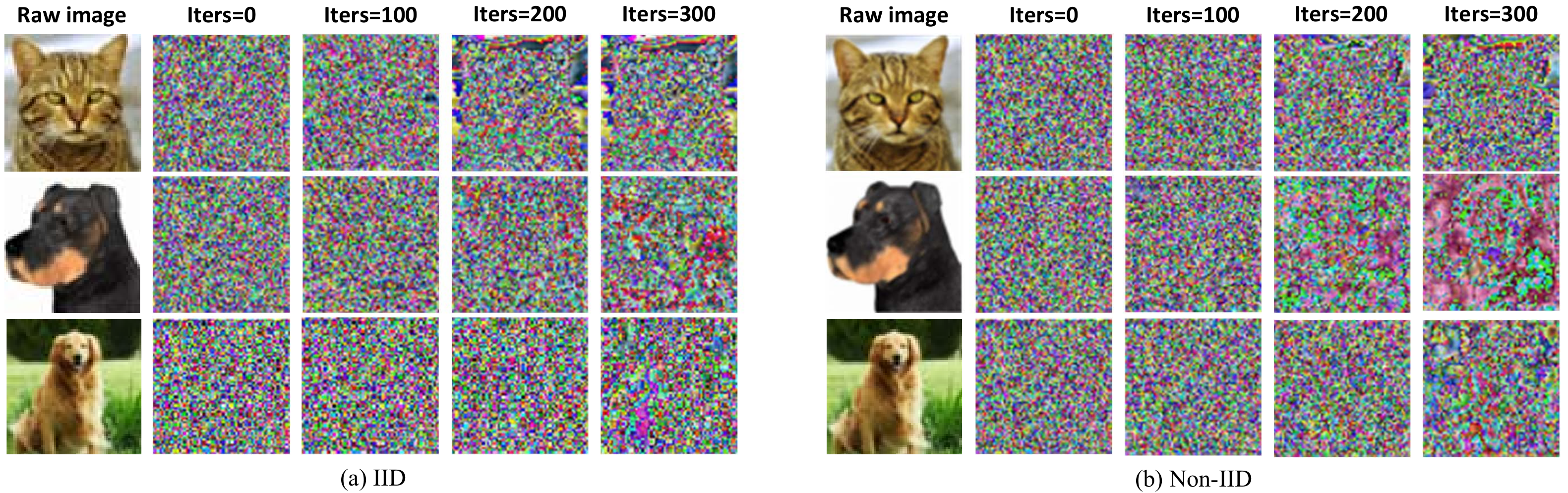}
  \caption{DLG attack against ADP with ratio of 0.25 on Animals.}
  \label{fig:attackanimal}
\end{figure*}

Tables \ref{table1} and \ref{table2} illustrate the comparison results of different methods under different ratio of injected noise to model parameters. First of all, it can be seen that as the amount of injected noise increases from 25\% to 75\%, the classification accuracy drops. This validates that there exists a trade-off between accuracy and privacy. Besides, we can observe that our ADP has a higher accuracy than the DP and Random DP on Blood dataset while achieving comparable accuracy to the other datasets. It suggests that the proposed ADP can improve the classification accuracy to some extent under privacy protection. 

\subsection{Impact of ADP on interpretability}
Next, we explore the influence of our ADP on the interpretability of classification results using three benchmark datasets. Figures~\ref{fig:ex-iid} and~\ref{fig:non-iid} illustrate the visual explanation of different methods using heatmap. We first take a look at the visualizations on MNIST. We can observe that DP no longer localizes the critical parts of handwritten numbers, leading to poor interpretability. In contrast, our ADP still has the ability to localize the most identifiable parts of each number, indicating that it can achieve higher interpretability and improve the recognition ability. In addition, the localizations of random DP are not as good as our method. We will further compare and explain their visualizations on the Blood dataset.

As shown in Figure~\ref{fig:ex-iid}(b), we can observe from the heatmap that the proposed ADP can better localize the salient regions of feature map compared to Random DP and No DP under IID data. It means that our method helps localize the lesions more precisely in medical examinations, thereby facilitating the understanding of the predicted results by doctors and patients. It thus can promote the deployment of FL in safety-critical applications.

% Here we want to that it focuses on blood cells, while the location focused on by With DP has shifted, especially on Non-IID data, and even in some images the location has been completely removed from the cells. Compared to Random DP, our method enables the model to focus more on cells and their interior, which means that our method helps to localize the lesions more precisely in medical examinations, facilitating the understanding of the predicted results by doctors and patients, and enhancing their trust. 

Next, we compare the visual interpretation of our method and the baselines on more complex Animals dataset. We can see from Figure~\ref{fig:ex-iid} (c) that our proposed method can capture the whole facial features and body features of animals, which is in line with human visual understanding. In contrast, No DP can hardly localize facial features and body features of animals while Random DP can only capture part of farcical features. We can conclude that our method can better explain the classification results on complex data.

% The interpretability of each model on IID data and Non-IID data are shown in Fig. \ref{fig:ex-iid} and \ref{fig:non-iid}, respectively. The comparison shows that the interpretability on Non-IID data is generally worse than that on IID data with the same settings, which is the same reason for the difference in accuracy.
What is more, we illustrate the visualizations of heatmap for different methods under the Non-IID scenario, as illustrated in Figure~\ref{fig:non-iid}. Compared to IID, all the methods will reduce their ability to localize the salient regions of feature map. Nevertheless, we can still observe from Figure~\ref{fig:non-iid} that our method can better interpret the prediction results.

The above extensive experiments illustrate that our proposed method can not only enhance interpretability but also improve classification accuracy. Therefore, the proposed ADP can help improve the trade-off between privacy, accuracy, and interpretability. 

% as it can be seen that the enhanced interpretability will allow the models to focus more on the key regions that have a great impact on the model classification. In summary, the enhanced interpretability benefits from a greater focus on the critical regions that have a strong influence on the classification performance.

\subsection{Robust to gradient leakage attack}

% \begin{figure*}[htbp] %通栏    跨行一排
% \begin{minipage}[t]{0.2\linewidth} %调节两个子图左右间距
% \centering
% \includegraphics[width=1.5in, height=1.3in]{pics/attack-25iidblood.pdf} %调节单个子图大小
% \caption{SOP} %子图下标题
% \label{fig:attack25iidblood} %引用标签
% \end{minipage}%
% \begin{minipage}[t]{0.25\linewidth}
% \centering
% \includegraphics[width=1.5in, height=1.3in]{pics/attack-25noniidblood.pdf}
% \caption{ESC}
% \label{fig:attack25noniidblood}
% \end{minipage}
% \end{figure*}

Finally, we also  show the robustness of the proposed ADP to gradient leakage attack. We adopt the well-known deep leakage from gradients (DLG) algorithm~\cite{Zhu2019deep} to attack our method. We want to verify whether the DLG method can reconstruct the input data based on the shared weights from local clients. Figures~\ref{fig:attackblood} and~\ref{fig:attackanimal} show the reconstruction of input data by the attacker under different training iterations. We can observe that DLG fails to recover the input data from the share weights under ADP. The main reason is that DLG is not very effective when the noise is large as mentioned in~\cite{Zhu2019deep}. In the proposed ADP, we add sufficient Gaussian noise with a variance ranging from $10^{-4}$ to $10^{-3}$ to the model parameters, which matches the range of noise scale to defend against gradient leakage attack as discussed in the prior work. Based on these experimental results, we can conclude that the proposed ADP is robust to the gradient leakage attack, thus is able to protect private data in federated learning.

% According to the parameter settings in the original paper, our data on Blood and Animals are very solid, so we purposely adjust the learning rate to improve its attack effect, but even so, we still have difficulty to be successfully attacked, and the results are shown in Figure 4, which provides a guarantee for our Adaptive DP to enhance the trade-off between privacy protection and interpretability.

% The test results on MNIST meet our expectation and are acceptable. This is because the model accuracy is still high after noise addition, and grayscale images with simple backgrounds are naturally easier to attack, while images with complex backgrounds and rich image elements are difficult to be successfully attacked, as the authors mentioned in their paper. Most importantly, the real-life data are all complex images, and the results on Animals demonstrate the privacy-preserving power of our approach. Furthermore, the reason why our approach was not successfully attacked on Blood and Animals may be that differential privacy provides sufficient privacy guarantees and the computed Gaussian noise variance is between $10^{-4}$ and $10^{-3}$, which is in line with the range of noise scale that needs to be added to render the attack ineffective as elaborated by the authors.

\section{Related Work}\label{sec:relatedwork}
We review the related work on privacy-preserving federated learning and the trade-offs between privacy protection, accuracy, and interpretability.
\subsection{Privacy-preserving federated learning}
In order to protect the sensitive information of users, researchers adopt different privacy-protecting methods to prevent gradient leakage attacks (GLK). These approaches can be categorized into two main groups: secure multiparty computing (SMC)~\cite{kanagavelu2020two,knott2021crypten,aono2017privacy} and differential privacy (DP)~\cite{Dwork2014}. SMC mainly adopts cryptographic techniques to protect the private information of input, making attackers hard to see the model updates. For instance, Kanagavelu et al.~\cite{kanagavelu2020two} proposed a two-phase SMC method to protect data privacy in federated learning. However, SMC is computational expensive, limiting its application to energy-aware IoT devices. DP aims to inject noise to the model parameters to defend against GLK. Wei et al.~\cite{Wei2020} proposed a privacy-protecting federated learning framework that adds noisy perturbation to the gradients of local clients before sending them to the server. While DP is very simple yet effective method for privacy protection, it will lead to an accuracy drop of federated learning~\cite{Dayan2021}.

% Federated learning has been widely used in many safety and privacy-critical applications, such as healthcare and finance. However
% Federated learning is a collaborative learning paradigm that carries out efficient machine learning through a server and multiple clients in a distributed manner, with training data stored locally~\cite{McMahan2017}. Due to its superior performance in protecting data privacy, federated learning is increasingly applied in various fields, especially in healthcare where highly sensitive medical data requires strong privacy protection~\cite{Kairouz2021,Rieke2020,Xu2021}. 

% Nevertheless, federated learning still faces privacy and security challenges. The updated parameters of the training model are vulnerable to malicious attacks, such as reconstruction~\cite{Melis2019} and inversion~\cite{Jeon2021}, when communicating between clients and the server, resulting in the leakage of sensitive information. Therefore, further privacy preserving mechanisms, such as differential privacy, secure multiparty computing, homomorphic encryption, are essential to prevent the privacy leakage of FL. Differential privacy is based on noise perturbation, not cryptography, and is widely applicable in deep learning models~\cite{Dwork2014,Abadi2016}. A differential privacy-based FL framework was proposed in~\cite{Wei2020}, where \textcolor{red}{artificial noise is added to parameters at the clients' side before model aggregation}, and the convergence behavior of FL with privacy-preserving noise perturbations was analyzed theoretically.

\subsection{Trade-off between privacy and accuracy}
Some studies~\cite{Harder2020,Bietti2022} try to improve the trade-off between privacy and accuracy. For example, Luo et al.~\cite{Luo2021} combined transfer learning and sparse network finetuning to improve the privacy-utility trade-off in federated learning. Hu et al.~\cite{Hu2022} developed Fed-SMP, a new differentially private FL framework for ensuring the client-level DP while retaining the model accuracy. In addition, some researchers~\cite{Kairouz2021,Agarwal2021} integrated secure aggregation and distributed DP to improve privacy-accuracy trade-off. These existing works can achieve remarkable performance on the trade-offs between privacy and model accuracy, they do not take into account the interpretability. In real-world applications, such as healthcare, it is crucial to explain and understand the diagnosis results of disease using federated learning~\cite{Meng2022}.

\subsection{Trade-off between interpretability and privacy}
Some recent works focus on the trade-off between accuracy and interpretability in classical machine learning (not federated learning)~\cite{Ribeiro2016,Pillai2021,Meng2022}. This is because trustworthy machine learning system needs to ensure both the privacy protection and interpretability. Harder et al.~\cite{Harder2020} attempted to produce accurate prediction while interpreting the classification results using several locally linear maps (LLM) per class. However, the proposed approach tries to explain the results based on input features using some simple datasets. Patel~et al.~\cite{Patel2022} designed a privacy-preserving framework to study the minimum privacy budget required for feature-based model explanations. In summary, existing works focus on the trade-off between privacy and \textit{feature-based interpretation rather} than gradient-based interpretation in regular machine learning. None of existing research works has explored the trade-off between privacy preservation and gradient-based interpretation in federated learning. To our best knowledge, the proposed adaptive differential privacy mechanism in this paper is the first work to fill the gap between them in federated learning.

\section{Conclusion}
This paper developed a simple yet effective adaptive differential privacy (ADP) method to overcome the trade-off between privacy and model interpretability. Specifically, we selectively inject noise into the model parameters based on the importance weights in the feature map. Evaluation results on multiple benchmark datasets illustrated that the proposed ADP can not only improve model interpretability but also guarantee privacy protection in the context of federated learning. In addition, our method is resistant to the well-known gradient leakage attack.

% Update the cvpr.cls to do the following automatically.
% For this citation style, keep multiple citations in numerical (not
% chronological) order, so prefer \cite{Alpher03,Alpher02,Authors14} to
% \cite{Alpher02,Alpher03,Authors14}.

% \begin{figure*}
%   \centering
%   \begin{subfigure}{0.68\linewidth}
%     \fbox{\rule{0pt}{2in} \rule{.9\linewidth}{0pt}}
%     \caption{An example of a subfigure.}
%     \label{fig:short-a}
%   \end{subfigure}
%   \hfill
%   \begin{subfigure}{0.28\linewidth}
%     \fbox{\rule{0pt}{2in} \rule{.9\linewidth}{0pt}}
%     \caption{Another example of a subfigure.}
%     \label{fig:short-b}
%   \end{subfigure}
%   \caption{Example of a short caption, which should be centered.}
%   \label{fig:short}
% \end{figure*}

%------------------------------------------------------------------------

%%%%%%%%% REFERENCES
{\small
\bibliographystyle{ieee_fullname}
\bibliography{reference}
}

\end{document}